\documentclass[english]{article}
\usepackage{babel,xpatch}
\usepackage[style=english]{csquotes}
\usepackage[style=ieee, backend=biber, url=false, doi=false]{biblatex}

\usepackage{geometry}
\usepackage{libertine}
\usepackage{hyperref}
\hypersetup{
	colorlinks = true,
	citecolor = {black},
	linkcolor = black
}

\usepackage{graphics} 
\usepackage{epsfig} 
\usepackage{amsmath} 
\usepackage{amssymb}  
\usepackage{bm}
\usepackage{subfig}
\usepackage{array}
\usepackage{algorithm, algorithmic}
\usepackage{placeins}

\title{\LARGE \bf
Minimum directed information:\\ A design principle for compliant robots
}

\author{Kevin Haninger\thanks{Corresponding email: \url{kevin.haninger@ipk.fraunhofer.de}. Author affiliated with the Fraunhofer Institut f{\"u}r Produktionsanlagen und Konstruktionstechnik (IPK).}%
}
\date{}

\begin{document}

\maketitle  

\begin{abstract}
A robot's dynamics -- especially the degree and location of compliance -- can significantly affect performance and control complexity. Passive dynamics can be designed with good regions of attraction or limit cycles for a specific task, but achieving flexibility on a range of tasks requires co-design of control. This paper takes an information perspective: the robot dynamics should reduce the amount of information required for a controller to achieve a threshold of performance in a range of tasks. Towards this goal, an iterative method is proposed to minimize the directed information from state to control on discrete-time nonlinear systems. iLQG is used to find a controller and value of information, then the design parameters of the dynamics (e.g. stiffness of end-effector or joint) are optimized to reduce directed information while maintaining a minimum bound on performance.  The approach is validated in simulation, on a two-mass system in contact with an uncertain wall position and a high-DOF door opening task, and shown to improve noise robustness and reduce time variance of control gains.
\end{abstract}

\section{Introduction}
Robotic tasks in semi-structured environments, such as contact-rich manipulation and human-robot interaction, are partially observed and the robot must respond appropriately to the current environment state. In physical tasks, this can be realized with intrinsic dynamics or active control, and the environment state either directly affects the robot dynamics, or is inferred and informs the control policy. 

Such tasks are often viewed from the impedance control framework \cite{hogan1984}, where the desired response to external forces or velocities is explicitly designed and realized through control. To adapt to a new task, the impedance can be changed by optimizing the impedance control parameters \cite{dimeas2015a, balatti2018, luo2019}, or impedance control can be used with other higher-level control, where it can improve reinforcement learning efficiency \cite{martin-martin2019} and robustness \cite{bogdanovic2019}. However, the rendered impedance depends heavily on the intrinsic robot dynamics (e.g. stiffness of end-effector/joints, reflected motor inertia), especially in dynamic events like collision, which have frequency content beyond the control bandwidth \cite{haddadin2010}.

So what design objectives should be applied to the intrinsic robot dynamics? Often, a low impedance is desired, which is a design goal in haptic devices \cite{massie1994}, human-robot interaction \cite{labrecque2017}, soft robotics \cite{albu-schaffer2008}, and soft actuators \cite{pratt1995}. However, reducing the intrinsic impedance comes with design trade-offs.  Classically, a lower stiffness reduces motion control bandwidth and accuracy \cite{bicchi2004}. The modelling and motion control of soft robots remains challenging, even with data-driven approaches \cite{beckers2019}. Additionally, compliance reduces the proprioceptive information flow - the ability to infer environment state or contact dynamics from robot position and force signals: intrinsic stiffness affects torque sensing accuracy \cite{kashiri2017} and the ability to infer contact conditions \cite{haninger2018}. 

A possible solution to this trade-off is variable impedance actuators, which can adapt their physical dynamics, e.g. the DLR David \cite{grebenstein2012}, but they currently present prohibitive cost and complexity.

\begin{figure}[h]
	\centering
	\includegraphics[width=.4\columnwidth]{ 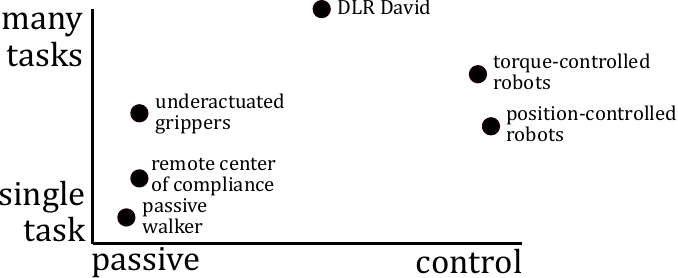} 
	\protect\caption{Range of capabilities of different hardware systems, and the degree to which their capabilities arise from passive dynamics or active control. \label{passive_dyn_vs_ctrl}}
\end{figure}

The intrinsic dynamics (i.e. dynamics without control) are important, and for certain tasks they are sufficient -- passive devices are established for locomotion \cite{mcgeer1990} and peg-in-hole assembly \cite{watson1978}. These systems are classically understood from a dynamics perspective; that their dynamics achieve stable limit cycles and good convergence properties. However, as illustrated in Figure \ref{passive_dyn_vs_ctrl}, purely passive devices do not have the adaptability which defines modern robotics. 

An alternative approach to the design of intrinsic dynamics is morphological computation \cite{muller2017}, which use information theoretic metrics to characterize the degree of influence between environment/robot \cite{ghazi-zahedi2017}, predictability \cite{bialek2001}, empowerment of a system \cite{klyubin2005}, or the influence of an external system \cite{williams2011}.  While information-theoretic approaches can scale to more complex systems described by data, they often do not include a sense of performance (objective/cost/reward function), thus limiting their applicability to design. The selection of minimal sensors to meet performance requirements has been considered on networked control systems \cite{tzoumas2020}, but not the co-design of dynamics and sensing.

This paper adopts an information-limited control approach, and proposes that good intrinsic dynamics reduce the amount of information required by the controller to achieve a certain performance in a range of tasks. In the limiting case of no online information required, the system is informationally passive, and can be operated open-loop (i.e. pure trajectory planning) \cite{massey1990, tanaka2017}. We propose that if only a small amount of information is required for control, suitable controllers may tolerate noise better or be more easily found by exploration. 

Existing results on linear systems are illustrative: the information rate required for a controller to stabilize a discrete-time linear system is $\sum\lambda_i^+$ bits/sample, where $\lambda^+_i$ are the unstable eigenvalues where $|\lambda_i|>1$ -- the more unstable, the more information required \cite{tatikonda2004}. Recent results have provided bounds on information rates required for linear time-varying systems with quadratic costs (i.e. LQG systems) \cite{tanaka2017}. These rates are measured with directed information \cite{massey1990}, also known as transfer entropy \cite{williams2011}, which has also been used to find bounds on the bits/sample required for control \cite{kostina2019}. 

\begin{figure}[h]
	\centering
	\includegraphics[width=.4\columnwidth]{ 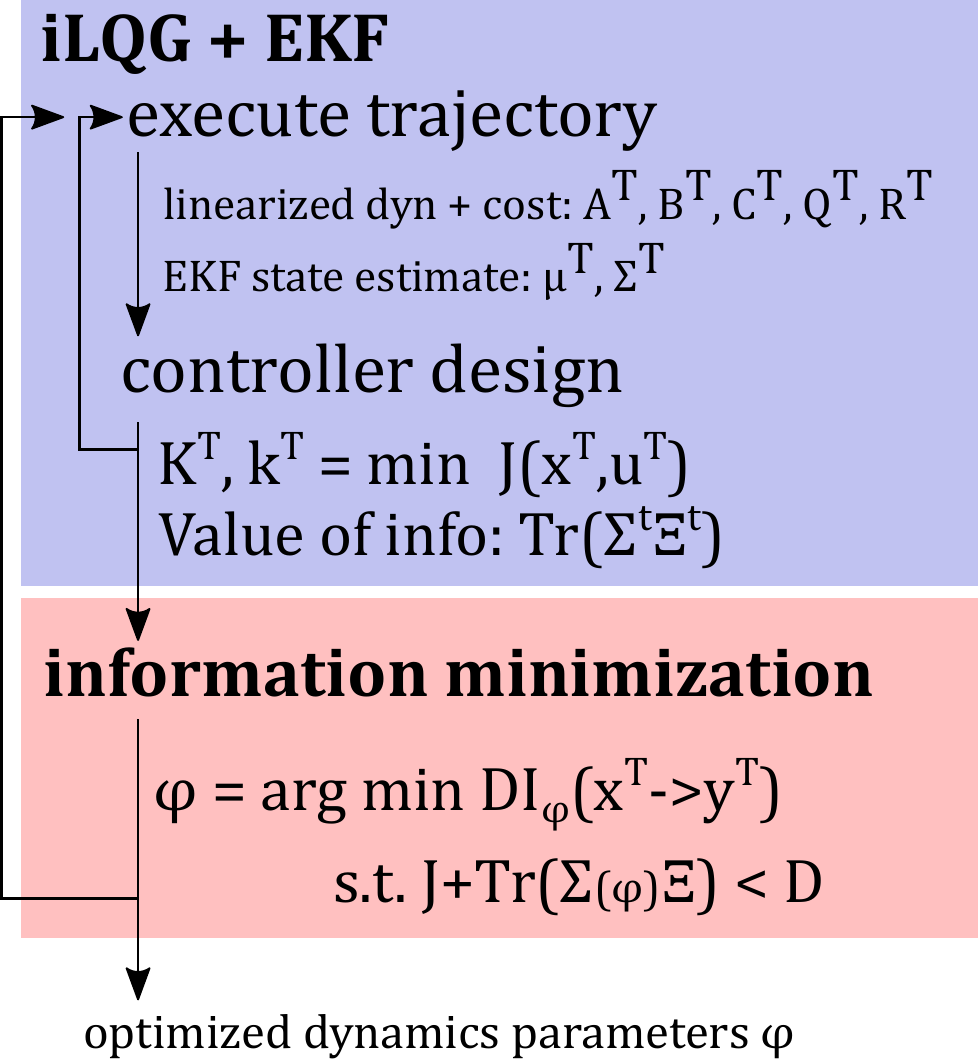} 
	\protect\caption{Proposed algorithm for minimizing directed information while maintaining a performance bound $D$. The result is optimized dynamics parameters $\phi$, such as end-effector or joint stiffness, link geometry, or center of compliance. \label{algorithm}}
\end{figure}

Here, an iterative method is proposed for finding intrinsic dynamic parameters which reduce the required directed information, as seen in Figure \ref{algorithm}.  About an initial trajectory, iLQG is used to find control, nominal performance, and the value of information \cite{todorov2005, tassa2012}. Parameters of the intrinsic dynamics are then updated to reduce the directed information while maintaining a bound on performance. The approach is validated in simulation, a two-mass system which makes contact with an uncertain environment position and a high-DOF door opening task, and shown to reduce sensitivity to environment noise. 

\subsection{Notation}
Denote a matrix determinant $|\cdot|$ and transpose $\cdot '$. Time series are denoted $y^t = [y_0, \dots, y_t]$.  A random variable is typeset $\mathrm{y}$ which takes values/samples $y$. Expectation is denoted $\mathbb{E}$, independence with $\perp$, and $\mathrm{y}\sim\mathcal{N}(\mu,\Sigma)$ denotes that $\mathrm{y}$ is distributed normally with mean $\mu$, covariance $\Sigma$.  

\section{Information and Performance}
\subsection{Directed Information}
Denote robot design variables as $\phi\in\Phi$, which may affect dynamics and/or sensing (e.g. end-effector compliance, joint stiffness, geometry of linkages). Let the robot and environment state at time step $t$ be $x_t\in\mathbb{R}^n$ with control input $u_t\in\mathbb{R}^m$. The directional influence of the state trajectory distribution of length $T$, $\mathrm{x}^T$, on control trajectory $\mathrm{u}^T$ can be characterized by directed information \cite{massey1990}
\begin{eqnarray}
\mathrm{DI}(\mathrm{x}^T\rightarrow \mathrm{u}^T) \triangleq \sum_{t=0}^T \mathrm{I}\left(\mathrm{x}^t;\mathrm{u}_t | \mathrm{u}^{t-1}\right), \label{DI_definition}
\end{eqnarray}
where conditional mutual information is
\begin{equation} 
 \mathrm{I}(\mathrm{x}^t;\mathrm{u}_t | \mathrm{u}^{t\mathtt{-}1}) \triangleq \\ \int p(x^t,u^t)\log\frac{p\left(x^{t},u_{t}|u^{t\mathtt{-}1}\right)}{p\left(x^{t}|u^{t\mathtt{-}1}\right) p\left(u_{t}|u^{t\mathtt{-}1}\right)}dx^tu^t.
\end{equation} 
As mutual information is non-negative and zero when the arguments are independent, $\mathrm{DI}(\mathrm{x}^T\rightarrow \mathrm{u}^T)$ is zero if and only if $\mathrm{u}_t$ is independent of $\mathrm{x}^t$, conditioned on $\mathrm{u}^{t-1}$, for all time. This occurs, for example, when there is no feedback and the system is operating in open-loop \cite{massey1990}. Reducing \eqref{DI_definition} means the control input is (in summation) more weakly related with the state. We propose that if acceptable performance can be reached with lower \eqref{DI_definition}, the control is simpler: more controller noise can be tolerated, and the state need not be so accurately estimated. 

\subsection{Dynamics and Performance}  
To formalize performance, consider the nonlinear coupled robot/environment dynamics as
\begin{equation}
\label{nl_dynamics}
\begin{aligned}
\mathrm{x}_{t+1} & =  f\left(\mathrm{x}_{t},\mathrm{u}_{t}\right)+\mathrm{w}_t  \\
\mathrm{y}_t & = g\left(\mathrm{x}_t\right) + \mathrm{v}_t
\end{aligned}
\end{equation}
where process and measurement noise are independently distributed as $\mathrm{w}_t \sim \mathcal{N}(0,W_t)$, $\mathrm{v}_t \sim \mathcal{N}(0,V_t)$, and the initial state as $\mathrm{x}_0\sim\mathcal{N}(0,X_0)$.  

Take a performance metric and objective of
\begin{align}
J(\mathrm{x}^{T},\mathrm{u}^{T})& =\mathbb{E}\left[\sum_{t=0}^{T-1}\ell_t(\mathrm{x}_t,\mathrm{u}_t)+\ell_T(\mathrm{x}_T)\right] \label{perf_metric}\\
J(\mathrm{x}^T,\mathrm{u}^T)& \leq D, \label{perf_bound} 
\end{align}
respectively, for some given performance threshold $D$.  The bound \eqref{perf_bound} captures that many robotic task outcomes are binary -- success/failure -- and the performance need not be optimal but rather meet some threshold for success. 

\subsection{Problem Statement}
Note that $\mathrm{DI}(x^T\rightarrow u^T)$ depends on the control policy, which may not be known at design time.  By Theorem 5 of \cite{derpich2013}, when $\mathrm{v}_t \perp \mathrm{w}_t$ and $(\mathrm{v}_t, \mathrm{w}_t) \perp \mathrm{x}_0$, and the control policy $\pi$ is deterministic given past observations $u_t=\pi(y^t)$,
\begin{eqnarray}
\mathrm{DI}(\mathrm{x}^T\rightarrow \mathrm{u}^T) \leq \mathrm{DI}(\mathrm{x}^T\rightarrow \mathrm{y}^T).
\end{eqnarray}
$\mathrm{DI}(\mathrm{x}^T\rightarrow \mathrm{y}^T)$ can be evaluated from dynamics and observation models alone, and will thus be used here. When the control policy is not deterministic, e.g. for exploration in reinforcement learning, additive control noise can considered with system noise $\mathrm{w}_t$. 

Let $J^i$ denote a performance metric of task $i$ and $D^i$ the corresponding minimum required performance which must be met by a control policy $u_t=\pi^i(y^t)$.  Then

\begin{equation}
\label{objective}
\begin{aligned}
\phi  = & \arg \min_{\phi \in \Phi} \mathrm{DI}_\phi(\mathrm{x}^T\rightarrow \mathrm{y}^T) \\  
&\,\, \mathrm{s.t.} \,\,\exists \pi^i, \,\, J^i_\phi(\mathrm{x}^T,\pi^i(\mathrm{y}^T)) \leq D^i \,\, \forall i \in I. 
\end{aligned}
\end{equation}
When the cost function of a performance metric $\ell_t$ is quadratic in state and dynamics are linear time-invariant (LTI) with Gaussian i.i.d. noise (i.e. LQG), the minimum information rate required for the controller to meet performance $D$ is  $R \leq \mathrm{DI}(\mathrm{x}^T\rightarrow \mathrm{u}^{T})$ \cite{tanaka2016}; where $R$ is an average number of bits/sample provided to the controller. Closed-form bounds on $R(D)$ have also been developed for LTI systems \cite{kostina2019}.

For linear time-varying systems, the selection of `virtual sensor' parameters and noise covariance to solve \eqref{objective} can be formulated as a semidefinite program \cite{tanaka2017}. The problem at hand is slightly different -- here, sensing is not freely chosen, but determined by design parameters, and these design parameters may also affect the dynamics. Furthermore; most problems of interest in robotics are nonlinear. Thus, here \eqref{objective} will be approached with gradients and barrier functions, to be introduced in the following two sections. 

\section{Iterative Linearization}
In the style of the iLQG algorithm \cite{tassa2012}, we consider approximations to \eqref{objective} based on local linearization about a previous trajectory. The iLQG is here extended to the partially observed case (i.e. iLQG + EKF), which allows approximation to the value of information. 

Given a trajectory $x^T, u^T$, consider state and control perturbations $\delta_x^T$, $\delta_u^T$, which approximate dynamics \eqref{nl_dynamics} as
\begin{equation}
\label{ltv_dynamics}
\begin{aligned}
\mathrm{x}_{t+1} & \approx f(x_t,u_t)+A_t \delta_{x,t} + B_t \delta_{u,t}+ \mathrm{w}_t \\
\mathrm{y}_t & \approx g(x_t) + C_t \delta_{x,t} + \mathrm{v}_t \\ 
\end{aligned}
\end{equation}
where $A_t=\frac{\partial f}{\partial x}$, $B_t=\frac{\partial f}{\partial u}$, $C_t = \frac{\partial g}{\partial x}$
 and all derivatives are evaluated at $x_t,u_t$. 

\subsection{Estimation and Directed Information}
Considering the directed information of the trajectory deviations, $\mathrm{DI}(\mathrm{\delta}_x^T\rightarrow \mathrm{y}^T)$ can be written largely following results for linear time-varying systems \cite{tanaka2017}, 
\begin{align}
\mathrm{I}(\mathrm{\delta}_x^t;\mathrm{y}_t | \mathrm{y}^{t-1}) & = \mathrm{H}(\mathrm{\delta}_x^t|\mathrm{y}^{t-1}) - \mathrm{H}(\mathrm{\delta}_x^t|\mathrm{y}^t) \nonumber\\
& = \mathrm{H}(\mathrm{\delta}_{x,t}|y^{t-1}) - \mathrm{H}(\mathrm{\delta}_{x,t}|y^t) \label{diff_ent}
\end{align}
with differential entropy $\mathrm{H}(\mathrm{x})\triangleq-\int p(x)\ln p(x)dx$, where \eqref{diff_ent} follows from the chain rule of entropy, conditional independence, and cancellation of terms \cite{cover1999}.  

Denoting belief $p(\delta_{x,t} |y^t) = \mathcal{N}(\mu_t,\Sigma_t)$, standard extended Kalman filter equations give
\begin{equation}
\Sigma_{t+1}  = ((A_t\Sigma_tA_t'+W_t)^{-1}+C_{t+1}'V_{t+1}^{-1}C_{t+1})^{-1} 
\end{equation}
where $\Sigma_0 = (X_0^{-1}+C_0'V_0^{-1}C_0)^{-1}$. The directed information can then be written as \cite{tanaka2017}
\begin{subequations}
	\begin{align}
    \mathrm{DI}(\mathrm{\delta}_x^T\rightarrow \mathrm{y}^T)& = \sum_{t=0}^T \frac{1}{2}\log|A_t\Sigma_tA_t'+ W_t|- \frac{1}{2}\log |\Sigma_t| \label{DI_1}\\
	& = \sum_{t=0}^T \frac{1}{2}\log|\Sigma_t^{\texttt{-}1}+ A_t'W_t^{\texttt{-}1}A_t|-\frac{1}{2}\log|W_t|. \label{DI_2}
	\end{align}
\end{subequations}

\subsection{Partially Observed iLQG}
To consider the performance constraint in \eqref{objective}, we extend the iLQG derivation in \cite{tassa2012} with additive noise and partial observations according to \eqref{ltv_dynamics}. Recall optimal cost to go, 
\begin{align}
J_{t}^{*}\left(\mathrm{x}_t\right)= \min_{u_{t}}\mathbb{E}\left[\ell_{t}\left(\mathrm{x}_{t},u_{t}\right)+J_{t+1}^{*}\left(\mathrm{x}_{t+1}\right)|\mathrm{y}^{t}\right] \label{bellman}
\end{align}
where $J_{T}^{*}(\mathrm{x}_T)=\mathbb{E}\,\ell_{T}(\mathrm{x}_T)$. Approximating the argument of \eqref{bellman} about perturbed state and input of $\tilde{\mathrm{x}}_t=\mathrm{x}_{t}+\delta_{x,t}$, $\tilde{u}=u_{t}+\delta_{u,t}$ and supressing subscript $t$ where unambiguous:
\begin{align*}
\ell_{t}\left(\tilde{\mathrm{x}},\tilde{u}\right) & \approx\ell_{t}\left(\mathrm{x},u\right)\texttt{+}\ell_{x}\delta_{x}\texttt{+}\ell_{u}\delta_{u}\texttt{+}\delta_{x}^{'}\ell_{xu}\delta_{u}\texttt{+}\frac{1}{2}\delta_{x}^{'}\ell_{xx}\delta_{x}\\
J_{t+1}^{*}\left(\mathrm{x}_{t+1}\right) & \approx J_{t+1}^{*}\left(f\left(\mathrm{x}_{t},u_{t}\right)\right)+\mathbb{E}\left[J_{x}^{+}\delta_{x+}+\frac{1}{2}\delta_{x+}^{'}J_{xx}^{+}\delta_{x+}\right]
\end{align*}
where $\delta_{x+}=A_t\delta_{x,t}+B_t\delta_{u,t}+\mathrm{w}_{t}$, $J_{x}^{+}=\frac{dJ_{t+1}^{*}}{d\mathrm{x}_{t+1}}$, $J_{xx}^{+}=\frac{d^{2}J_{t+1}^{*}}{d\mathrm{x}^{2}_{t+1}}$, and $\ell_x = \frac{\partial \ell_t}{\partial x_t}$. This then gives a change in the argument to \eqref{bellman} of
\begin{equation}
 \approx  \mathbb{E}\left[Q_{x}\delta_{x}+Q_{u}\delta_{u}+ \frac{1}{2}\delta_{x}'Q_{xx}\delta_{x}+ \right. \\ \left. \delta_{x}'Q_{ux}'\delta_{u}+\frac{1}{2}\delta_{u}'Q_{uu}\delta_{u}|\mathrm{y}^{t}\right] +\frac{1}{2}\mathrm{Tr}(J_{xx}^{+}W_{t})
 \label{approx_bellman}
\end{equation}
where $Q_{x}=\ell_{x}+J_{x}^{+}A_t$, $Q_{u}=\ell_{u}+J_{x}^{+}B_t$, $Q_{xx}=\ell_{xx}+A'_tJ_{xx}^{+}A_t$, $Q_{uu}=\ell_{uu}+B_t'J_{xx}^{+}B_t$, $Q_{ux}=\ell_{ux}+B'_tJ_{xx}^{+}A_t$. 

If $p(\delta_x | \mathrm{y}^t) = \mathcal{N}(\mu_t, \Sigma_t)$, \eqref{approx_bellman} is minimized by 
\begin{equation}
\delta u^{*}=-Q_{uu}^{-1}\left(Q_{u}+Q_{ux}\mu_t\right)=k_t+K_t\mu_t \label{optimal_controller}
\end{equation} which gives  
\begin{equation}
J_t^*\left(\tilde{x}_{t}\right) \approx \ell_t(x,u) + J_{t+1}^*(f(x_t,u_t)) +\Delta J_{t}+J_{x}\mu_t+\mu_t 'J_{xx}\mu_t\\ + \frac{1}{2}\mathrm{Tr}(\Xi_t \Sigma_t) + \frac{1}{2}\mathrm{Tr}(J_{xx}^+ W_t) \label{exp_cost}
\end{equation}
where $\Delta J_t = k'Q_{uu}k+k'Q_{uu}$, $J_{x}=Q_{x}+K'Q_{uu}k+K'Q_u + Q_{ux}'k$, $J_{xx}=Q_{xx}+K'Q_{uu}K+K'Q_{ux}+Q_{ux}'K$, and $\Xi_t = Q_{xu}Q_{uu}^{-1}Q_{ux}$. Note that $\Xi_t = K_t'(\ell_{uu}+B_t'J_{xx}^+B_t)K_t$, and $\mathrm{Tr}(\Xi_t\Sigma_t)$ is the new cost introduced by belief uncertainty (i.e. the value of information). 

\subsection{Differences with respect to fully-observed iLQG}
In comparison the standard iLQG, there are two additional terms in \eqref{exp_cost} arising from process noise covariance $W_t$ and belief uncertainty $\Sigma_t$. 

One important additional distinction is that the state is not known, so the point to linearize the dynamics about is also uncertain. We adopt the heuristic of maintaining a moving average for reference trajectories $\overline{x}^T$ and $\overline{u}^T$. For a new trajectory, the maximum likelihood state estimate $\mu^T$ is used as $  \overline{x}^T  \leftarrow \alpha \mu^T + (1-\alpha)\overline{x}^T$ and $\overline{u}^T  \leftarrow \alpha u^T + (1-\alpha)\overline{u}^T$ for some $\alpha\in  [0,1]$. 

Furthermore, the cost of an individual trajectory is not deterministic, making the use of cost reduction to schedule regularization \cite{tassa2012} or as a terminating condition more difficult. 

\section{Optimizing information}

To avoid higher-order tensors or vectorization, consider a single element of the design parameters $\phi_{i}\in\mathbb{R}$ which affects both the observations $g_{\phi}\left(x_{t}\right)$ and/or the dynamics $f_{\phi}\left(x,u\right)$. Changing this parameter can affect the directed information $\mathrm{DI}\left(\mathrm{x}^{T}\rightarrow \mathrm{y}^{T}\right)$, the belief covariance $\Sigma_{t}$, and performance $J^*_t$. This section optimizes \eqref{objective} by gradient descent, including the constraint via barrier methods. 

\subsection{Objective}
Let precision matrix $\Gamma_t=\Sigma_t^{-1}$ and denote $\nabla_i\Gamma_{t}=\frac{\partial\Gamma_{t}}{\partial\phi_{i}}$ , $\nabla_i A_{t}=\frac{\partial^{2}f}{\partial x\partial\phi_{i}}|_{x_{t},u_{t}}$ and $\nabla_i C_{t}=\frac{\partial^{2}f}{\partial u\partial\phi_{i}}|_{x_{t},u_{t}}$. Calculating the gradient of the terms in the objective in the form of \eqref{DI_1}:
\begin{equation}
\frac{\partial}{\partial\phi_{i}}\mathrm{DI}(\delta_x^T\rightarrow y^T)  = \\ \sum_{t=0}^T\frac{1}{2}\mathrm{Tr}\left(\left(A_t'\Sigma_tA_t+W_t\right)^{-1}\Upsilon_{t,i}\right) +\frac{1}{2}\mathrm{Tr}\left(\nabla_i\Gamma_t\Sigma_t\right) \label{obj_grad}
\end{equation}
where $\Upsilon_{t,i} = \nabla_iA_t'\Sigma_tA_t+ A_t'\Sigma_t\nabla_i\Gamma_t\Sigma_tA_t+A_t'\Sigma_t\nabla_iA_t$.

This is found by standard matrix relations \cite{petersen2008}, including $\frac{d\Sigma_{t}}{d\phi_{i}}=\Sigma_{t}\nabla_i\Gamma_t\Sigma_{t}$. Calculating the gradient via \eqref{DI_2} was found to be less numerically stable.

$\nabla_i\Gamma_{t}$ can be calculated recursively as
\begin{equation}
\nabla_i\Gamma_{t+1}= N_0 - N_1 + N' \nabla_i\Gamma_{t}N \label{d_gamma}
\end{equation}
where $N_0 = \nabla_iC_{t}'V_{t}^{-1}C_{t}+C_{t}'V_{t}^{-1}\nabla_i C_{t}$, $N_1 = \Psi_t\left(\nabla_i A_{t}\Sigma_{t}A_{t}'+A_{t}\Sigma_t\nabla_i A_{t}'\right)\Psi_t$,  $N =\Psi_t A_{t}\Sigma_{t}$, and  $\Psi_t=\left(A_{t}\Sigma_{t}A_{t}'+W_{t}\right)^{-1}$. Recursive expressions such as \eqref{d_gamma} may not be stable. For a simplified stability analysis, consider time-invariant system matrices and derivatives, where a sufficient condition for the convergence of the sequence, i.e. $\nabla \Gamma_{t+1} = \nabla \Gamma_{t}$, is if all eigenvalues of $N$ are stable $|\lambda_j(N)|<1  \forall j$, arising from the existence condition for the discrete-time Lyapunov equation \cite{davis2010}.

\subsection{Performance constraint}

The performance constraint can then be incorporated by barrier methods, where 
\begin{align*}
\phi^* = \arg\min & \,\,\mathrm{DI}\left(\phi\right)  \,\,\mathrm{s.t.}\,\,J_0^{*}(\phi)\leq D\\
\approx\arg\min &\,\, \mathrm{DI}\left(\phi\right)-\beta\log\left(D-J_0^{*}(\phi)\right)
\end{align*}
where $\beta>0$ is a scalar weight for the barrier penalty. The gradient of the barrier can be found as
\begin{align*}
\frac{\partial}{\partial\phi_{i}}\left[-\beta\log\left(D-J_{0}^{*}\right)\right] & = -\frac{\beta}{D-J_{0}^{*}}\frac{\partial J_0^*}{\partial\phi_{i}}.
\end{align*}

We simplify by considering only the impact of $\phi$ on belief covariance $\Sigma_t$, leaving the impact of $\phi$ on dynamics to be handled with the control optimization in the iLQR, giving
\begin{eqnarray}
\frac{\partial J_0^* }{\partial \phi_i }&\approx\sum_{t=1}^{T}\mathrm{Tr}\left(\Xi_t\Sigma_{t}\frac{d\Gamma_{t}}{d\phi_{i}}\Sigma_{t}\right) \label{perf_grad}
\end{eqnarray}
Recall $\Xi_t$ is the value of information from \eqref{exp_cost}. 

\subsection{Gradient Descent}
Simple gradient descent is used to update $\phi$ as
\begin{align}
   \phi & \leftarrow \phi + h \left(\frac{\partial}{\partial \phi} \mathrm{DI} - \frac{\beta}{D-J_0^*} \frac{\partial J_0^*}{\partial\phi}\right) \label{grad_step}\\ 
   h & = \mathrm{tanh}(D-J_0^*+\gamma)
\end{align}
where gradients are given in \eqref{obj_grad} and \eqref{perf_grad}, $J_0^*$ is taken from the iLQG solution, small positive constant $\gamma$, and the step size $h$ reduces as the cost barrier is approached.

\section{Validation}
To validate the approach, an implementation\footnote{Implementation available at \url{https://github.com/khaninger/info_min}} is built based on the DRAKE robotics library \cite{tedrake2019}. Two example systems are tested: a two-mass system with rigid contact, and a simplified door-opening task with a high-DOF robot. Optimizing compliance is shown to improve noise robustness and reduce the time variance in iLQR gains, which has been proposed as a metric for control complexity \cite{ruckert2013}.
\begin{figure}
	\centering
	\subfloat[\label{two-mass-model}]{\includegraphics[width=.6\columnwidth]{ 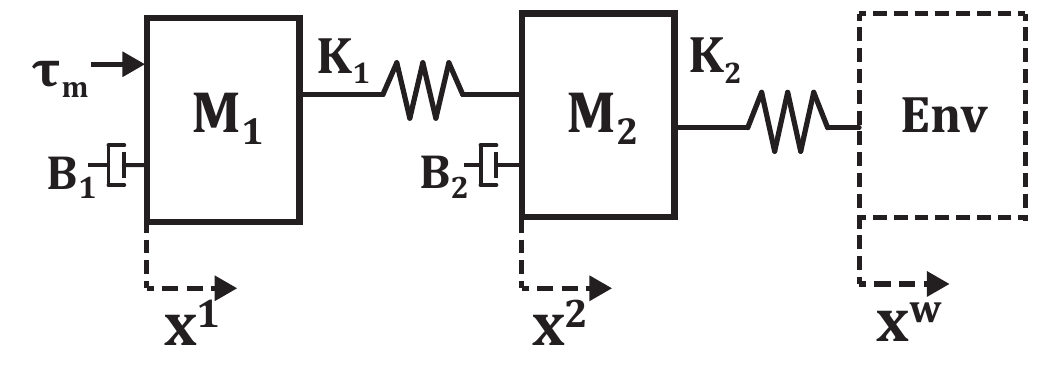}}
	\hfill
	\subfloat[\label{door-opening}]{\includegraphics[width=.35\columnwidth]{ 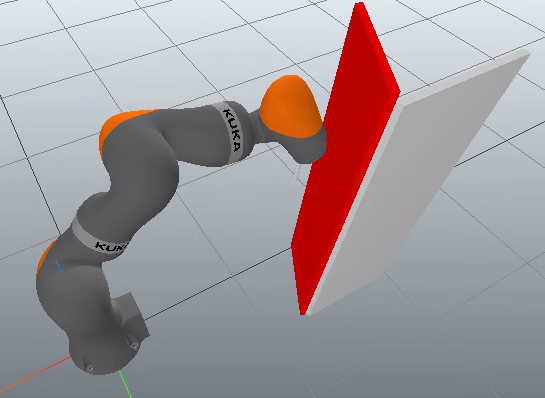}}
	\protect\caption{In (a), a two mass system in contact (in free-space when $x^2<x^w$, $K_2=0$). In (b), a door opening simulation, robot EE is coupled with a diagonal stiffness to door handle.}
\end{figure}

\begin{figure}
	\centering
	\subfloat[ Total cost and directed information (bits) \label{two_mass_perf_and_DI}] {\includegraphics[width=.5\columnwidth]{ 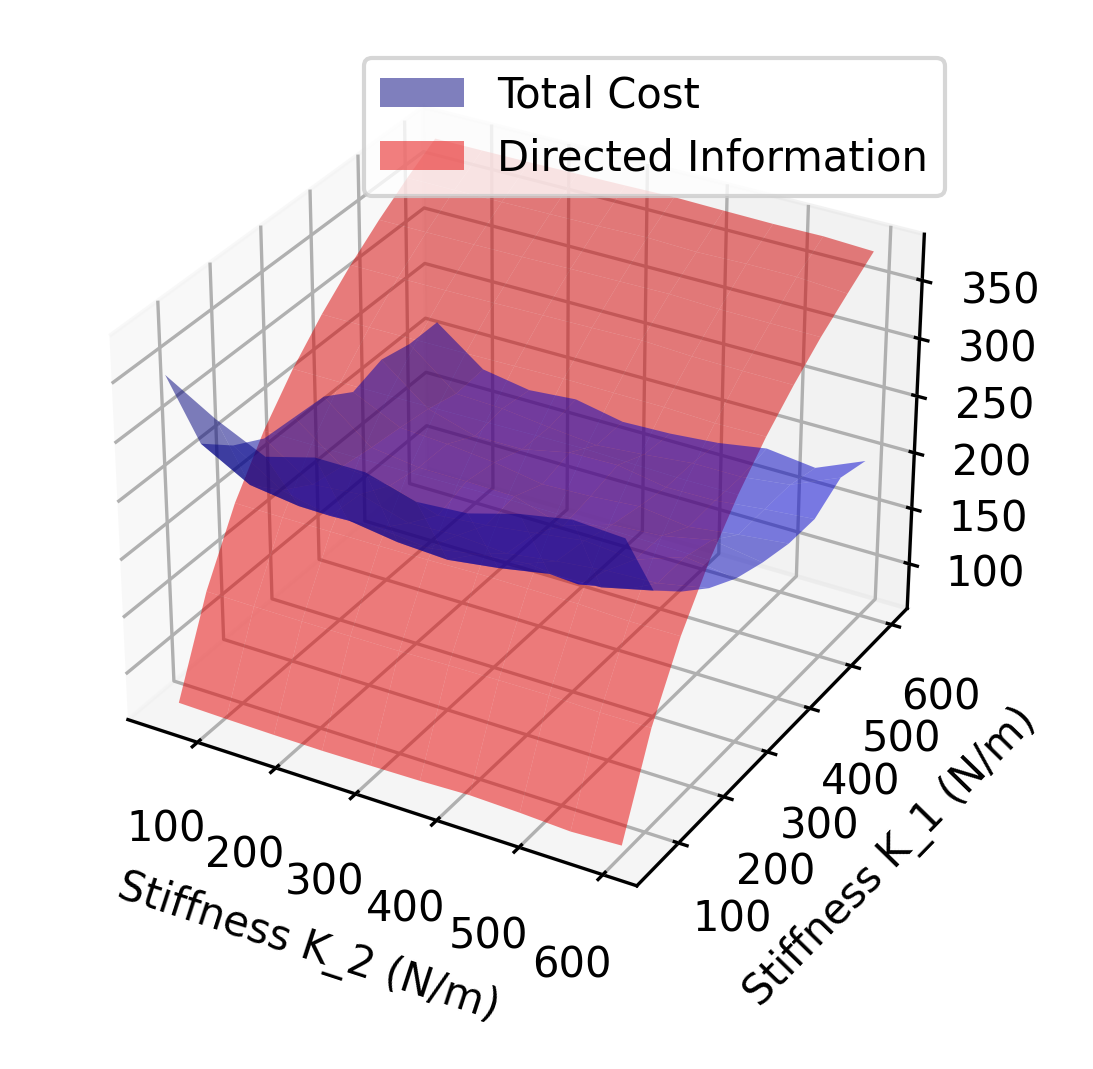}} 
	\hfill
	\subfloat[Trace of optimization over DI (top) and total cost (bottom)\label{di} \label{perf}]{\begin{tabular}[b]{c}
	\includegraphics[width=0.45\columnwidth]{ 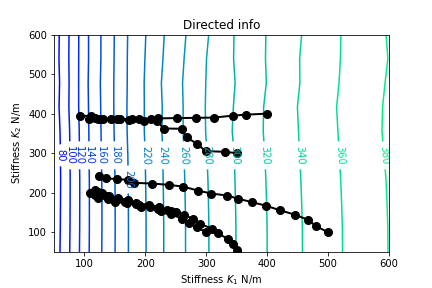}\\ 
	\includegraphics[width=0.45\columnwidth]{ 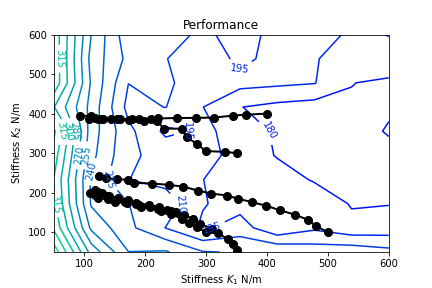}
	\end{tabular}}
	\caption{The parameter space and traces of the optimization}
\end{figure}

\subsection{Two-mass with contact}

A standard two-mass system is taken with inertia, damping, and stiffness of $M_i$, $B_i$, and $K_i$ for $i=1,2$, and positions of $x^1, x^2$. A rigid wall is added at $\mathrm{x}^w$, where $\mathrm{x}^w\sim p(x^w)$ varies between trajectory rollouts, but is fixed within the task. This system emulates a flexible-joint robot with joint-torque sensing coming in contact with a rigid environment of uncertain location. The stiffness $K_1$ is the joint stiffness, and $K_2$ the effective stiffness of end-effector and environment in series. The bounds $K_1 \in [50, 600]$ (N/m) and $K_2 \in [100, 600]$ (N/m) will be applied, with fixed parameters $M_\cdot = 0.8, 0.4$ Kg and $B_\cdot = 60, 35$ Ns/m, for $_1$, $_2$, respectively.
\begin{figure}[h!]
	\centering
	\subfloat[iLQG result, $K_1 = 500 N/m$, $K_2 = 500 N/m$ \label{orig}]{\includegraphics[width=0.5\columnwidth]{ 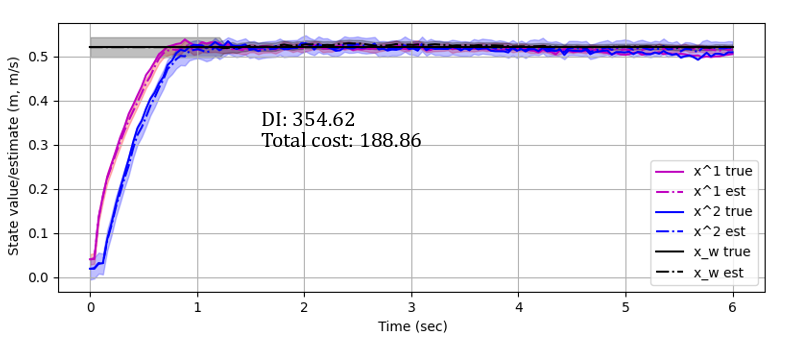}}\\ 
	\vspace{-10pt}
	\subfloat[iLQG result, $K_1 = 123 N/m$, $K_2 = 240 N/m$ \label{optimized}] {\includegraphics[width=0.5\columnwidth]{ 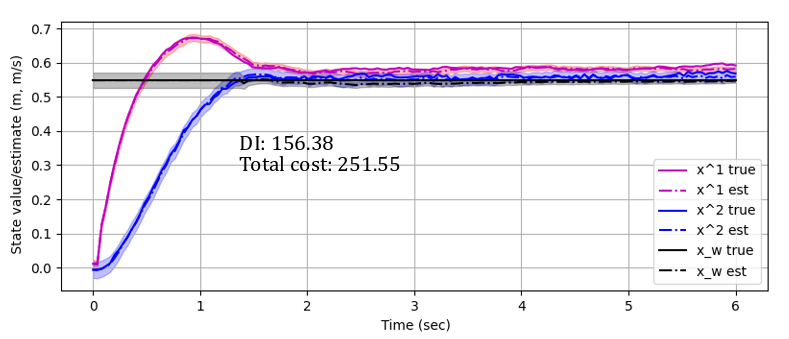}} 
	\caption{Plots of true state, mean and covariance of corresponding belief, for initial and optimized gains. Uncertainty in the state estimate is increased; but acceptable performance is maintained.}
\end{figure}
\begin{figure}
	\centering
	\includegraphics[width=.7\columnwidth]{ 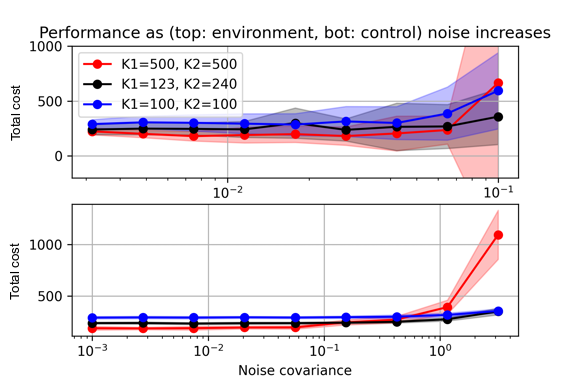}
	\caption{iLQG total cost (mean and covariance over 20 trajectories) as noise is increased in (top) environment ($x_w$ position) and control (additive Gaussian noise). The optimized stiffness has better noise robustness than the high-stiffness system.\label{noise_robustness}}
\end{figure}
The measurements are position and force $y_t = [x^1; K_1(x^2-x^1)]+v_t$ where $v_t \sim \mathcal{N}(0,\mathtt{diag}(1e-3, 1e-2)])$. The dynamics are standard, with the caveat that when $x^2 < x^w$, $K_2$ is effectively $0$. Process noise has covariance $W=\mathtt{diag}(1e\mathtt{-}7, 5e\mathtt{-}3, 1e\mathtt{-}7, 1e\mathtt{-}2, 0)$ and initial state covariance is $X0 = 10^{-3}\mathtt{diag}(1, 1, 1, 1, 0.1)$ with state of $[x^1, \dot{x}^1, x^2, \dot{x}^2, x^w]$. The objective is to achieve a small positive contact force $f^o$ with the environment (which is a basic requirement for contact tasks), 
\begin{equation*}
\ell_t = \Vert(K_2(x^2-x^w)_+)-f^o\Vert+0.1\Vert\dot{x}^1\Vert+0.1\Vert\dot{x}^2s\Vert +10^{-4}\Vert\tau_m\Vert
\end{equation*}
where $(x)_+$ is  $\frac{1}{2}(\sqrt(x^2+\gamma^2)+x)$, the differentiable approximation to $\max(0,x)$ from \cite{tassa2012}. Here, $f_0 = 10$, $\gamma = 0.1$, and the barrier weight $\beta = 1e4$. 
\begin{figure*}
	\centering	\subfloat[Low stiffness {$k_{1:3} = [1,1,1]1e4$} ]{\includegraphics[width=0.33\textwidth]{ 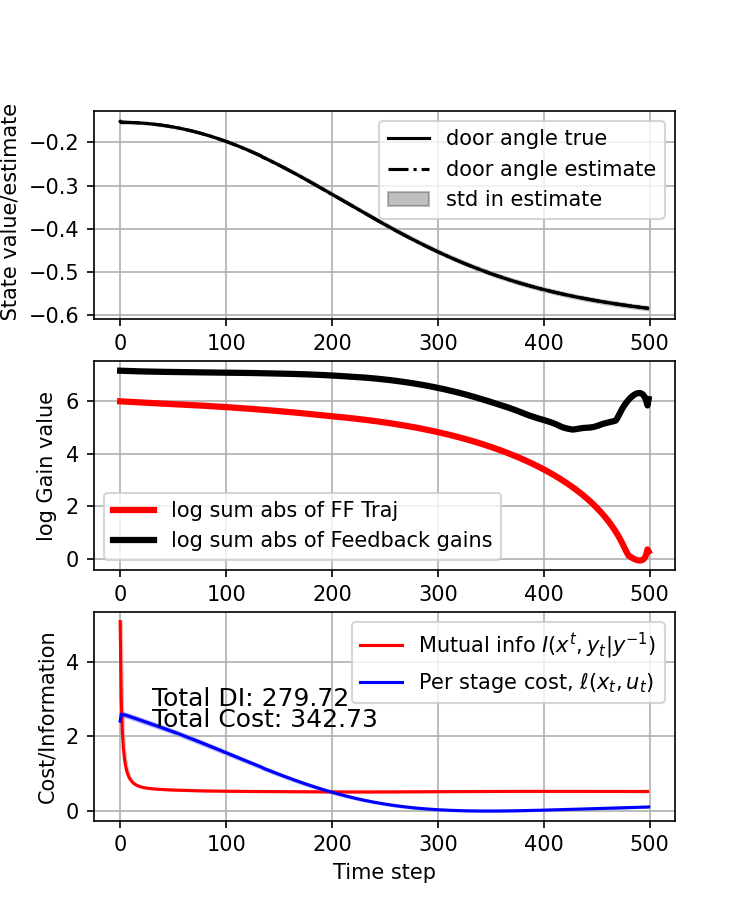}} 
	\subfloat[Optimized stiffness {$k_{1:3} = [9.8, 9.3, 7.9]e5$}]{\includegraphics[width=0.33\textwidth]{ 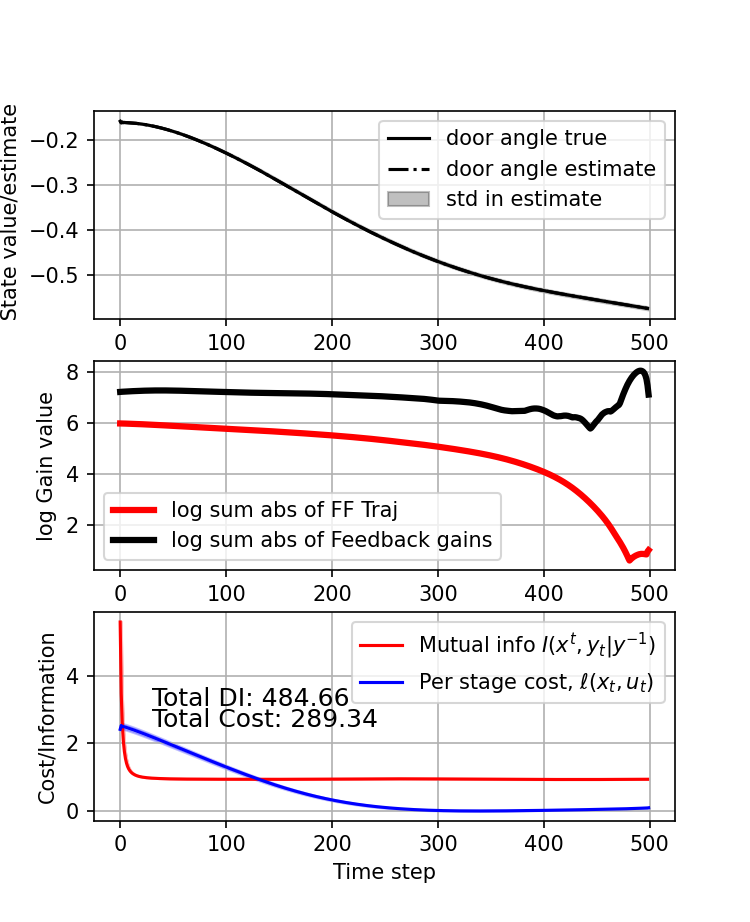}} 
	\subfloat[High stiffness {$k_{1:3} = [1,1,1]1e6$}]{\includegraphics[width=0.33\textwidth]{ 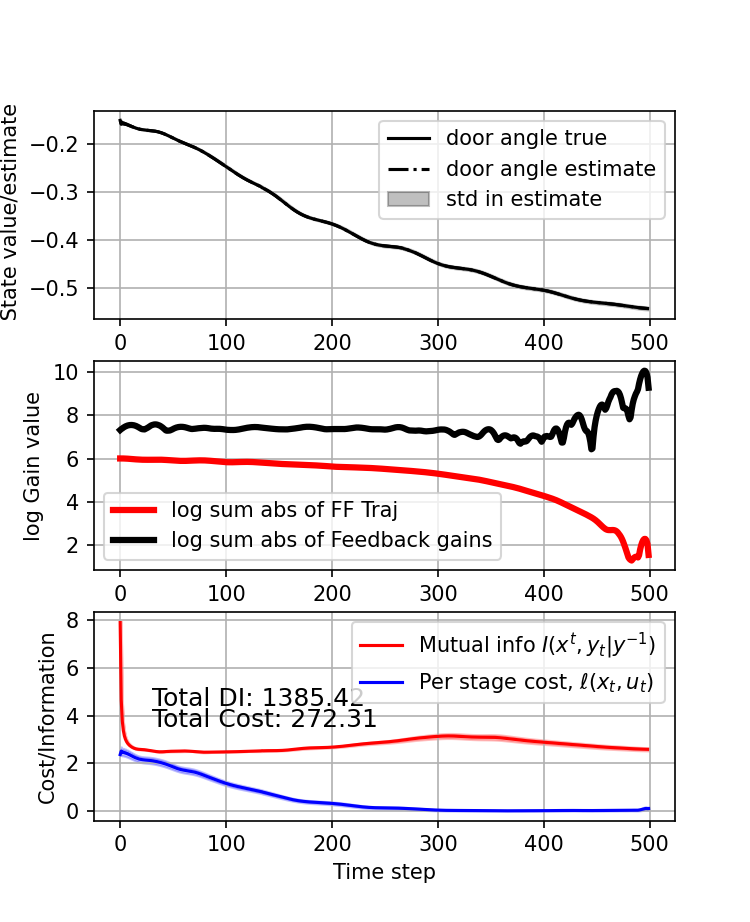}} 
	\caption{Optimization of the stiffness coupling robot to door reduces the temporal variability of the feedback gains compared with the stiff system, reducing directed information by almost 1/3 with only a $6\%$ penalty in total cost, relative to high-stiffness gains. \label{kuka}}
\end{figure*}
First, iLQG was performed over a grid of $K_1$ and $K_2$, and the directed information of the converged controller estimated. The total cost and DI can be seen in Figure \ref{two_mass_perf_and_DI}. The total cost decreases as $K_1$ or $K_2$ increases, and reducing $K_1$ reduces the directed information. 

The stiffnesses $\phi = {K_1, K_2}$ can be optimized according to \eqref{grad_step}, where a number of initial $\phi$ can be seen on contour plots of directed information \ref{di} and performance \ref{perf}, where $D = 250$. The barrier method succeeds in increasing $K_2$ from lower initial values, but the trajectories do not converge to a single point as the barrier gradient is almost parallel to the DI graddient. A trace of the system response with initial stiffnesses $K_1 = 500$, $K_2 = 500$ can be compared with the optimized stiffneses $K_1 = 123$, $K_2 = 240$ is shown in Figure \ref{orig} and \ref{optimized}, respectively. 

The noise robustness of several $\phi$ was compared: how the total cost of iLQG varies as either (1) environment or (2) control noise increases, where environment noise is the covariance of normally distributed $x_w$ and the control noise is additive Gaussian. The high stiffness system is more sensitive to all noise. The optimized gains show better environment noise robustness, even compared with the low-stiffness system. However, the optimized gains have slightly more control noise sensitivity than the low stiffnesses. This suggests that optimizing to reduce required information allows the system to better tolerate environment noise.

\subsection{Manipulator opening door}

To investigate the scalability of the algorithm, it is then applied to a simulation of the KUKA iiwa manipulator in a door opening/closing task.  The robot and door handle are coupled via a diagonal stiffness matrix $K_e = \mathtt{diag}[k_1, k_2, k_3, k_4, k_5, k_6]$ to simulate the compliance of a gripper. Torsional stiffnesses are fixed at $k_{4:6} = 50$ Nm/rad, and linear stiffnesses $\phi = k_{1:3}$ are optimized. The robot positions and velocities are observed, joint-space torque control is used, and the cost function is 
\begin{equation*}
	\ell_t = \Vert \theta - \theta_0 \Vert + 0.05 \Vert \dot{q} \Vert + 0.1 \Vert \tau_m \Vert
\end{equation*}
where $\theta$ is the angle of the door, and $\dot{q}$, $\tau_m$ the joint velocities and torque of the robot. This objective function aims to move the door to the position $\theta_0$, while reducing the velocity and motor torque along the way. 

Due to space limitations, interested readers are referred to \href{https://github.com/khaninger/info_min}{the public code} for parameter values used and the calculation of the required derivatives. A very low-stiffness $k_{1:3} = [1,1,1]1e4$ (N/m) and high-stiffness $k_{1:3} = [1,1,1]1e6$ coupling are compared with the optimized stiffnesses where $k_{1:3} = [9.8, 9.3, 7.9]e5$. A video is available at \url{https://youtu.be/Mp5HAMRCdAg} and the resulting state and control trajectories are seen in Figure \ref{kuka}. The iLQR feedback gains and feedforward trajectory vary more over time when the stiffness is higher, providing better performance from higher directed information.  The optimized gains respect the performance limit of $D=300$ while reducing DI. Similar trends occur in a door closing task, as seen in the attached video.

\section{Conclusion}
This paper proposed a new principle for the design of intrinsic dynamics: to reduce the information required for control, operationalized by iLQG where gradients of the linearized dynamics and value of information reduce the directed information while maintaining performance. The feasibility and stability of the algorithm was validated on a two-mass model; and a higher-DOF example used to demonstrate the scalability. Optimizing compliance was shown to reduce the time variation in control gains and the environment noise sensitivity.

\section*{Acknowledgements}
This project has received funding from the European Union's Horizon 2020 research and innovation programme under grant agreement No  820689 — SHERLOCK.  The author would also like to thank Raul Vicete-Garcia, Manuel Baum, and the reviewers for conversations which contributed to the paper development. 
\printbibliography
\end{document}